\pdfoutput=1

\documentclass[11pt]{article}

\usepackage[preprint]{acl}

\usepackage{times}
\usepackage{latexsym}
\usepackage{amsmath}
\usepackage[T1]{fontenc}

\usepackage[utf8]{inputenc}

\usepackage{microtype}

\usepackage{inconsolata}

\usepackage{graphicx}
\usepackage{cleveref}
\usepackage{booktabs}
\usepackage{multicol}
\usepackage{multirow}
\usepackage{adjustbox}
\usepackage{ulem}
\usepackage{color}
\usepackage{tabularx}

\title{Fact-Level Confidence Calibration and Self-Correction} %

\author{
\textbf{Yige Yuan}$^{1,2}$, \textbf{Bingbing Xu}$^{1}$, \textbf{Hexiang Tan}$^{1,2}$, \textbf{Fei Sun}$^{1}$, \\\textbf{Teng Xiao}$^{3}$, \textbf{Wei Li}$^{1}$, \textbf{Huawei Shen}$^{1,2}$, \textbf{Xueqi Cheng}$^{1,2}$ \\
$^{1}$Institute of Computing Technology, Chinese Academy of Sciences\\
$^{2}$University of Chinese Academy of Sciences, 
$^{3}$Penn State University
}

\begin{document}
\maketitle
\begin{abstract}
Confidence calibration in LLMs, i.e., aligning their self-assessed confidence with the actual accuracy of their responses, enabling them to self-evaluate the correctness of their outputs. However, current calibration methods for LLMs typically estimate two scalars to represent overall response confidence and correctness, which is inadequate for long-form generation where the response includes multiple atomic facts and may be partially confident and correct. These methods also overlook the relevance of each fact to the query.
To address these challenges, we propose a fact-level calibration framework that operates at a finer granularity, calibrating confidence to relevance-weighted correctness at the fact level. Furthermore, comprehensive analysis under the framework inspired the development of confidence-guided fact-level self-correction (\textbf{ConFix}), which uses high-confidence facts within a response as additional knowledge to improve low-confidence ones. Extensive experiments across four datasets and six models demonstrate that ConFix effectively mitigates hallucinations without requiring external knowledge sources such as retrieval systems\footnote{Code is available at \url{https://github.com/yuanyige/fact-calibration}}.
\end{abstract}
\section{Introduction}

Large Language Models (LLMs) have recently shown remarkable capabilities in understanding and generating language that closely resembles human communication~\cite{DBLP:conf/nips/BrownMRSKDNSSAA20,DBLP:journals/corr/abs-2303-08774}.
Nonetheless, a major obstacle to their reliability is the prevalence of hallucinations~\cite{lin2021truthfulqa, zhang2023siren, li2023halueval, golovneva2022roscoe, bang2023multitask}, a phenomenon where the models generate incorrect and unreliable outputs.
This issue not only undermines user trust but also restricts the application of LLMs in domains where reliability is crucial, such as in the legal, financial, and educational fields.

Echoing the ancient adage that \textit{``To know what you know and what you do not know, that is true wisdom''}, confidence calibration in LLMs emerges as an effective method to mitigate  hallucinations~\cite{li2024think,liu2023lightweight,huang2024calibrating}.
By confidence calibrating, models can better align their self-assessed confidence with the actual accuracy of responses, empowering them to self-evaluate the correctness of  outputs. This mechanism offers an effective way to identify hallucinations by using the model's confidence as a basis  to either trust or question the model's response.

\begin{figure}[t]
    \centering
    \includegraphics[width=\linewidth]{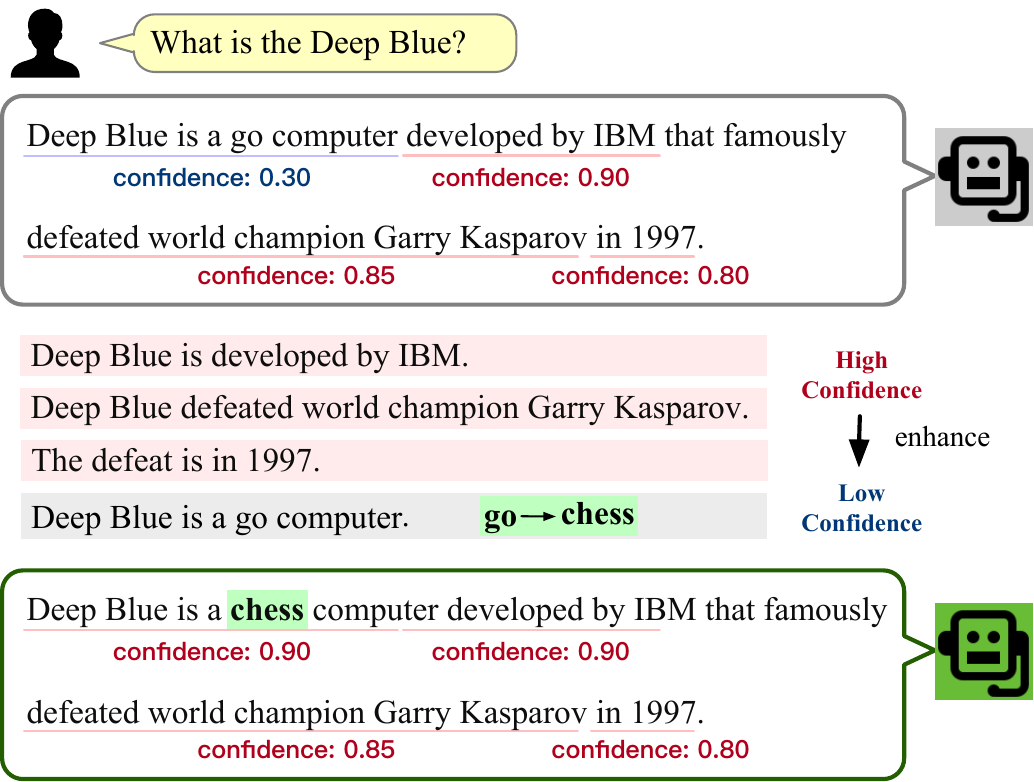}
    \vskip -0.5em
    \caption{Motivation of our fact-level confidence calibration and confidence-guided self-correction.}
    \label{fig:self-corr1}
\vskip -1.5em
\end{figure}

However, current confidence calibration methods for LLMs~\cite{guo2017calibration,DBLP:conf/emnlp/NguyenO15} typically estimate two scalars to represent the overall confidence and correctness for the entire response. This approach is unreasonable for long-form generation, where responses may contain multiple atomic facts (illustrated in~Fig.\ref{fig:self-corr1}). In such cases, considering the varying of facts in one response, the confidence and correctness should also be diverse, capable of reflecting higher certainty in some facts and greater uncertainty in others. 
Furthermore, within long-form responses~\cite{xiao2024cal,xiao2024leverage}, certain facts exist that may indeed be correct but lack relevance to the query. Previous calibration methods predominantly~\cite{huang2024calibrating,liu2023lightweight} focus on  correctness while neglecting the relevance.

To address these challenges, we propose a novel framework for confidence calibration operating at a finer fact-level granularity. Within this framework, the confidence assessment of each fact incorporates two key aspects: correctness and relevance. Correctness indicates the factual accuracy of the fact, while relevance measures the extent to which the fact is related to the query. Calibration of a response is defined as the degree of alignment between confidence and correctness weighted by relevance across all facts.
This framework endows the model with the capability to exhibit partial confidence and correctness in individual facts.
Extensive analysis based on the aforementioned framework yields three interesting findings: 
(1) Calibration on fact-level imposes a stricter standard than calibration on response-level.
(2) Overconfidence issue stems from implicit facts.
(3) High variance exists in fact-level confidence within a response.

The aforementioned three observations inspire the development of Confidence-Guided Fact-Level Self-Correction (\textbf{\texttt{ConFix}}) to enhance the generation and mitigate hallucinations (illustrated in~Fig.\ref{fig:self-corr1}). For a response, \texttt{ConFix} first leverages the aforementioned framework to segment the response into multiple facts and evaluate their confidence vector. 
It then uses the high-confidence facts and their associated confidence score as additional knowledge to augment low-confidence facts, with the aim of all facts within the response achieving high confidence. 
\texttt{ConFix} can self-enhance to mitigate hallucinations without the need for external knowledge sources such as retrieval systems. 
Experiments with \texttt{ConFix} across four datasets and six models reveal that it can significantly reduce the occurrence of hallucinations, thereby increasing the models' reliability and enabling their real-world  applications. \textbf{In summary, our main  contributions are}:

\noindent \textbf{Fact-Level Calibration Framework}: The proposed fact-level calibration framework operates at a finer level of granularity to align the confidence with the correctness weighted by relevance across all facts. This framework enables LLMs to exhibit partial confidence and accuracy in individual facts.

\noindent \textbf{Insightful Observations}: We uncover insightful observations regarding the model's scale and its calibration capability on fact- and response-levels. 

\noindent \textbf{Self-Correction Method}: We propose \texttt{ConFix} based on the fact-level calibration framework to enhance the generation and reduce hallucinations without relying on external knowledge sources.

\section{Related Work and Formulation}

\paragraph{Related Work} 
Several prior works, such as SelfCheckGPT~\cite{manakul-etal-2023-selfcheckgpt,wang2024fine,zhang2024luq} and so on~\cite{mundler2023self}, focus on uncertainty-based hallucination detection. 
While these methods share a common goal with ours, there are several key distinctions:
(1) Their focus is primarily on detection, whereas our work goes beyond that by investigating the mechanisms of fine-grained confidence estimation and identifying three novel phenomena.
(2) We introduce a self-correction method that functions without relying on external knowledge, in contrast to their approaches, which are limited to detection.
(3) While their methods operate at the sentence level, we take a more granular approach by breaking down text into atomic facts, enabling a deeper analysis of confidence at the fact level.
(4) Furthermore, we propose a general framework for confidence estimation, with their methods representing specific instances within this broader approach.
There are also works like FactScore~\cite{DBLP:journals/corr/abs-2305-14251} that focus on fact-level factuality evaluation by leveraging external knowledge. In contrast, our method centers on the model's internal confidence and calibration, providing a self-correction solution that operates without the  external knowledge.

\paragraph{Problem Formulation} Let the dataset be $D = \{\mathbf{x}_1, \mathbf{x}_2, \ldots, \mathbf{x}_N\}$, where $\mathbf{x}_i$ represents the $i$-th query, and the model's responses are $A = \{\mathbf{y}_1, \mathbf{y}_2, \ldots, \mathbf{y}_N\}$. 
Each query-answer pair $(\mathbf{x}_i, \mathbf{y}_i)$ is associated with a confidence score $\mathrm{conf}_i$, indicating the model’s certainty to the answer, and a correctness score $\mathrm{corr}_i$, measuring the objective accuracy of the answer. 
In long-form generation, we extend this to the fact level, where each response $\mathbf{y}_i$ contains $M_i$ individual facts $\{f_{i}^{j}\}_{j=1}^{M_i}$, with each fact $f_{i}^{j}$ assigned a relevance score $\mathrm{rel}_{i}^{j}$, a correctness score $\mathrm{corr}_{i}^{j}$, and a confidence score $\mathrm{conf}_{i}^{j}$. 
The aim is to align confidence with relevance-weighted correctness across all facts in each response.

\begin{figure*}[t]
    \centering
    \includegraphics[width=0.95\linewidth]{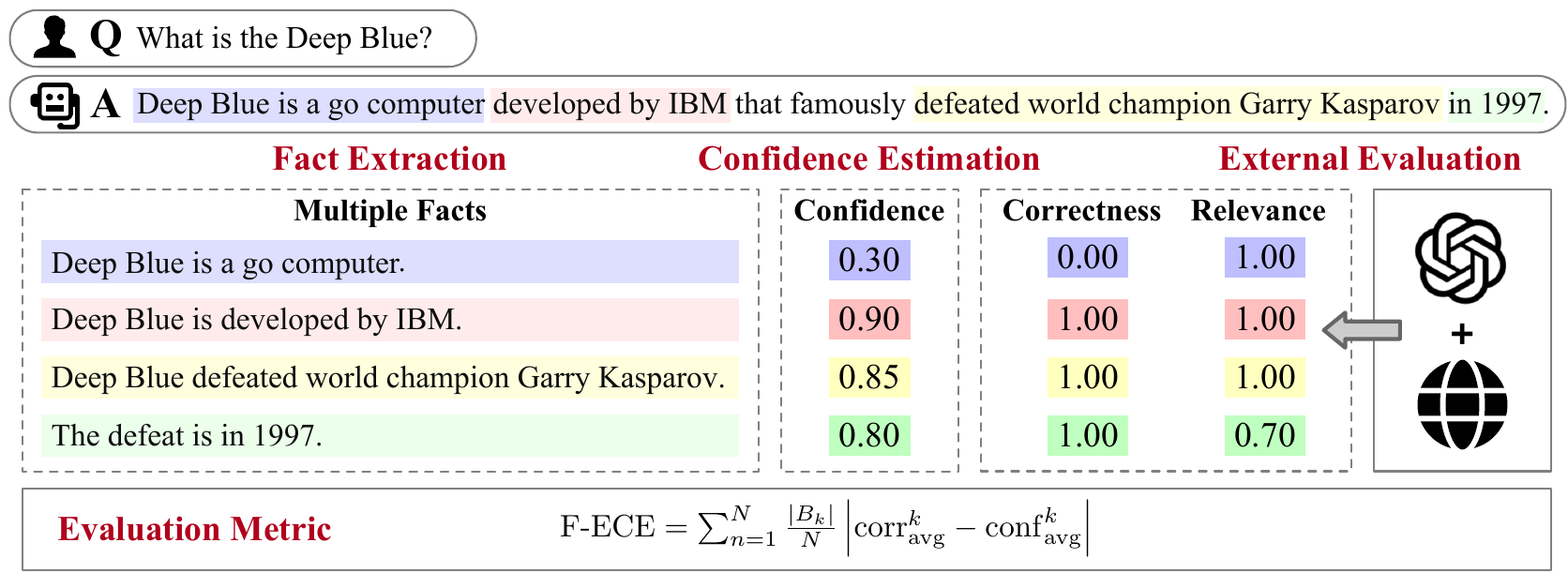}
    \vskip -0.5em
    \caption{An illustration of our fact-level confidence calibration framework for fine-grained LLM calibration.}
    \label{fig:fact-cal}
\end{figure*}

\section{Fact-Level Confidence Calibration}

In this section, we first outline the motivation and provide a overview of the Fact-Level Confidence Calibration framework. 
Following this, we discuss three key observations arising from our framework that offer deeper insights. 
Finally, we conclude by summarizing how these observations inform and shape our approach to conduct self-correction.

\subsection{Motivation}

Compared to the confidence calibration for short-form generation or traditional classification problems, a significant challenge in calibrating long-form text generation is that a response may contain multiple facts, making it unreasonable to assign a single correctness measure and a single confidence score to the entire response.
The reason is that the answer might be partially correct, and the model might also be partially confident in only a subset of the facts of a response.
Meanwhile, some facts in response are irrelevant to the query, so the calibration based solely on correctness is insufficient.

Based on the above motivation, our proposed calibration framework aims to calibrate the confidence to relevance-weighted correctness on the fact level, which leads to the following two advantages:
(1) \textbf{Finer Granularity}: 
we assign a confidence vector rather than a scalar to a response, where each item represents confidence for a single fact. This fine-grained framework allows for more nuanced and precise calibration.
(2) \textbf{Relevance Awareness}: 
we assess both the correctness and the relevance of each fact,
which ensures that the confidence score attributed to each fact can reflect its significance and appropriateness within the given context.

\begin{table*}[h]
\renewcommand{\arraystretch}{0.85}
\centering
\fontsize{9}{10.5}\selectfont\setlength{\tabcolsep}{0.47em}
\caption{Comparison of confidence estimation and calibration performance across five base models for three confidence estimation methods (Verbalization, isTrueLogit, Consistency) on the ASQA and LongFact datasets.}
\vskip -1em
\label{tab:ece}
\begin{adjustbox}{width=0.98\linewidth}
\setlength{\tabcolsep}{2mm}
\begin{tabular}{llcccccccccccc}
\toprule
\textbf{Dataset ($\rightarrow$)} &
& \multicolumn{4}{c}{\textbf{ASQA}} & \multicolumn{4}{c}{\textbf{LongFact}} \\
\cmidrule(lr){3-6} \cmidrule(lr){7-10}
\multicolumn{2}{l}{\textbf{Metric ($\rightarrow$) $\slash$ Model ($\downarrow$) }} 
& \multicolumn{2}{c}{\textbf{Confidence}} & \multicolumn{2}{c}{\textbf{Calibration}} & \multicolumn{2}{c}{\textbf{Confidence}} & \multicolumn{2}{c}{\textbf{Calibration}} \\
\cmidrule(lr){3-4} \cmidrule(lr){5-6} \cmidrule(lr){7-8} \cmidrule(lr){9-10}
\multicolumn{2}{l}{\textbf{Method ($\downarrow$) $\slash$ Level ($\rightarrow$)}}
& \textbf{Fact} & \textbf{Response} & \textbf{Fact} & \textbf{Response} & \textbf{Fact} & \textbf{Response} & \textbf{Fact} & \textbf{Response} \\
\midrule
\multirow{6}{*}{Verbalization} 
&  \texttt{Llama-2-7b} 
& 0.6946 & 0.7110 & 0.2605 & 0.2508 
& 0.6753 & 0.7842 & 0.2109 & 0.1415
\\ 
&  \texttt{Llama-2-13b}  
& 0.7629 & 0.8289 & 0.2404 & 0.2606 
& 0.7510 & 0.8418 & 0.1556  & 0.1310
\\ 
&  \texttt{Vicuna-7b} 
& 0.8054 & 0.8425 & 0.3365 & 0.3524 
& 0.7567 & 0.8591 & 0.1505 & 0.1369
\\ 
&  \texttt{Vicuna-13b} 
& 0.8661 & 0.9339 & 0.2540 & 0.2688 
& 0.8838 & 0.9299 & 0.1134 & 0.1094 
\\ 
&  \texttt{GPT-3.5-turbo} 
& 0.9620 & 0.9673 & 0.1789 & 0.1855 
& 0.9659 & 0.9554 & 0.0856 & 0.0935
\\ 
\cmidrule(lr){2-10}
& Comparison & 0.8182 & 0.8567 & 0.0040 &	0.0043 & 0.8065	& 0.8741 & 0.0025	& 0.0006\\
\midrule
\multirow{6}{*}{isTrueLogit} 
&  \texttt{Llama-2-7b} 
& 0.6639 & 0.8088 & 0.2972 & 0.3748 
& 0.5081 & 0.8624 & 0.3525 & 0.1998
\\ 
&  \texttt{Llama-2-13b}  
& 0.9741 & 0.9483 & 0.3723 & 0.3764 
& 0.8693 & 0.9696 & 0.1635 & 0.1932
\\ 
&  \texttt{Vicuna-7b} 
& 0.5448 & 0.5672 & 0.2867 & 0.2806 
& 0.5607 & 0.6186 & 0.2998 & 0.2539
\\ 
&  \texttt{Vicuna-13b} 
& 0.8116 & 0.6342 & 0.2486 & 0.2837 
& 0.8648 & 0.7947 & 0.1240 & 0.2213
\\ 
&  \texttt{GPT-3.5-turbo} & - & - & - & - & - & - & - & - \\ 
\cmidrule(lr){2-10}
& Comparison & 0.7486 & 0.7396 & 0.0018	& 0.0000 & 0.7007	& 0.8113 & 0.0167 &	0.0003\\
\midrule
\multirow{6}{*}{Consistency} 
&  \texttt{Llama-2-7b} 
& 0.8714 & 0.8334 & 0.3224 & 0.3331  
& 0.9276 & 0.9760 & 0.2167 & 0.1660
\\ 
&  \texttt{Llama-2-13b}  
& 0.8903 & 0.8320 & 0.3101 & 0.3326 
& 0.8940 & 0.8980 & 0.1385 & 0.1804
\\ 
&  \texttt{Vicuna-7b} 
& 0.8834 & 0.8227 & 0.3312 & 0.3373 
& 0.9397 & 0.9767 & 0.1946 & 0.1477
\\ 
&  \texttt{Vicuna-13b} w
& 0.9250 & 0.9007 & 0.2473 & 0.2500 
& 0.9727 & 0.9767 & 0.1020 & 0.1085
\\ 
&  \texttt{GPT-3.5-turbo} 
& 0.8859 & 0.8393 & 0.1572 & 0.1807 
& 0.9613 & 0.9480 & 0.0715 & 0.0957
\\ 
\cmidrule(lr){2-10}
& Comparison & 0.8912 &	0.8456 & 0.0080	& 0.0069 & 0.9391 &	0.9551 & 0.0047 &	0.0014\\
\bottomrule
\end{tabular}
\end{adjustbox}
\vspace{-0.1in}
\end{table*}

\subsection{Architecture}
\label{sec:cal_frame}
To calibrate confidence with relevance-weighted correctness at fact level, 
our framework comprises four components, as shown in~\cref{fig:fact-cal}: (1) fact extraction, (2) correctness and relevance evaluation, (3) confidence estimation, (4) calibration evaluation.

\paragraph{Fact Extraction} 
Given query-answer pair $(\mathbf{x}_i, \mathbf{y}_i)$, the first step is to extract the individual facts from response. 
This extraction can be efficiently carried out using a powerful external language model, such as GPT-4~\cite{DBLP:conf/nips/BrownMRSKDNSSAA20}, producing a set of facts $\{f_{i}^{j}\}_{j=1}^{M_i}$  for $\mathbf{y}_i$.

\paragraph{Correctness and Relevance Evaluation}
Once facts are extracted, we evaluates both the correctness and relevance of each fact in relation to the query. 
Correctness is assessed by verifying the factual accuracy of each fact using GPT-4 in combination with retrieval methods, such as search engines and ground truth answers from relevant datasets, yielding correctness scores $\{\mathrm{corr}_i^j\}_{j=1}^{M_i}$.
Similarly, relevance scores $\{\mathrm{rel}_i^j\}_{j=1}^{M_i}$ can also be measured using GPT-4 to capture each fact’s pertinence to the query within the context of the overall response.

\paragraph{Confidence Estimation}
Confidence estimation quantifies the certainty of the LLM for each fact in its response. 
To derive a confidence vector $\{\mathrm{conf}_{i}^{j}\}_{j=1}^{M_i}$ for each response, we employ three widely used confidence estimation methods:
(1) Verbalization-based method~\cite{tian2023just}: model is prompted to verbally express confidence score for each fact in the response.
(2) Logit-based method~\cite{DBLP:journals/corr/abs-2207-05221}: confidence is determined by the probability assigned to the ``True'' token when evaluating the response as True or False.
(3) Consistency-based method~\cite{manakul-etal-2023-selfcheckgpt}: confidence is determined by the consistency across multiple sampled outputs.
Formally, the confidence of $f_i^j$ can be represented as: 
\begin{align}
\mathrm{conf}_i^j = \mathcal{C}\left(\mathrm{LLM}(\cdot), p_c(f_i^j,\mathbf{x}_i,\mathbf{y}_i)\right), \end{align} 
where $\mathrm{LLM}$ is the model to be evaluated, $p_c$ is the prompt, $\mathcal{C}$ denotes the method used for estimation.

\paragraph{Calibration Evaluation}
\label{calibration metric}
We define Fact-Level Expected Calibration Error (F-ECE) as the evaluation metric used to quantify the discrepancy between a model's confidence and the relevance-weighted correctness across all responses and their corresponding facts. 
For each fact within a response, we calculate the relevance-weighted correctness as the product of the fact’s correctness score and its relevance score. 
The response-level relevance-weighted correctness and confidence are then computed by averaging these values across all facts within a response, as detailed in~\cref{eq:ece_1}.
\begin{equation}
\begin{split}
\mathrm{corr}_i &= \frac{1}{M_i} \sum_{j=1}^{M_i} \mathrm{corr}_i^{j} \times \mathrm{rel}_i^{j}, \\
\mathrm{conf}_i &= \frac{1}{M_i} \sum_{j=1}^{M_i} \mathrm{conf}_i^{j}, 
\label{eq:ece_1}
\end{split}
\end{equation}
F-ECE is then computed by assessing the difference between the average relevance-weighted correctness and the average confidence within each bin $k$, where $B$ is the number of bins for grouping confidence scores, and $B_k$ is the set of responses in the $k$-th bin. 
Let $\mathrm{corr}^{k}_{\mathrm{avg}}= \frac{1}{|B_k|} \sum_{i \in B_k} \mathrm{corr}_i$ and $\mathrm{conf}^{k}_{\mathrm{avg}}= \frac{1}{|B_k|} \sum_{i \in B_k} \mathrm{conf}_i$,
\begin{equation}
\mathrm{F\text{-}ECE} = \sum_{n=1}^{N} \frac{|B_k|}{N} \left| \mathrm{corr}^{k}_{\mathrm{avg}} - \mathrm{conf}^{k}_{\mathrm{avg}} \right|
\end{equation}

\begin{figure}[b]
    \centering
    \includegraphics[width=\linewidth]{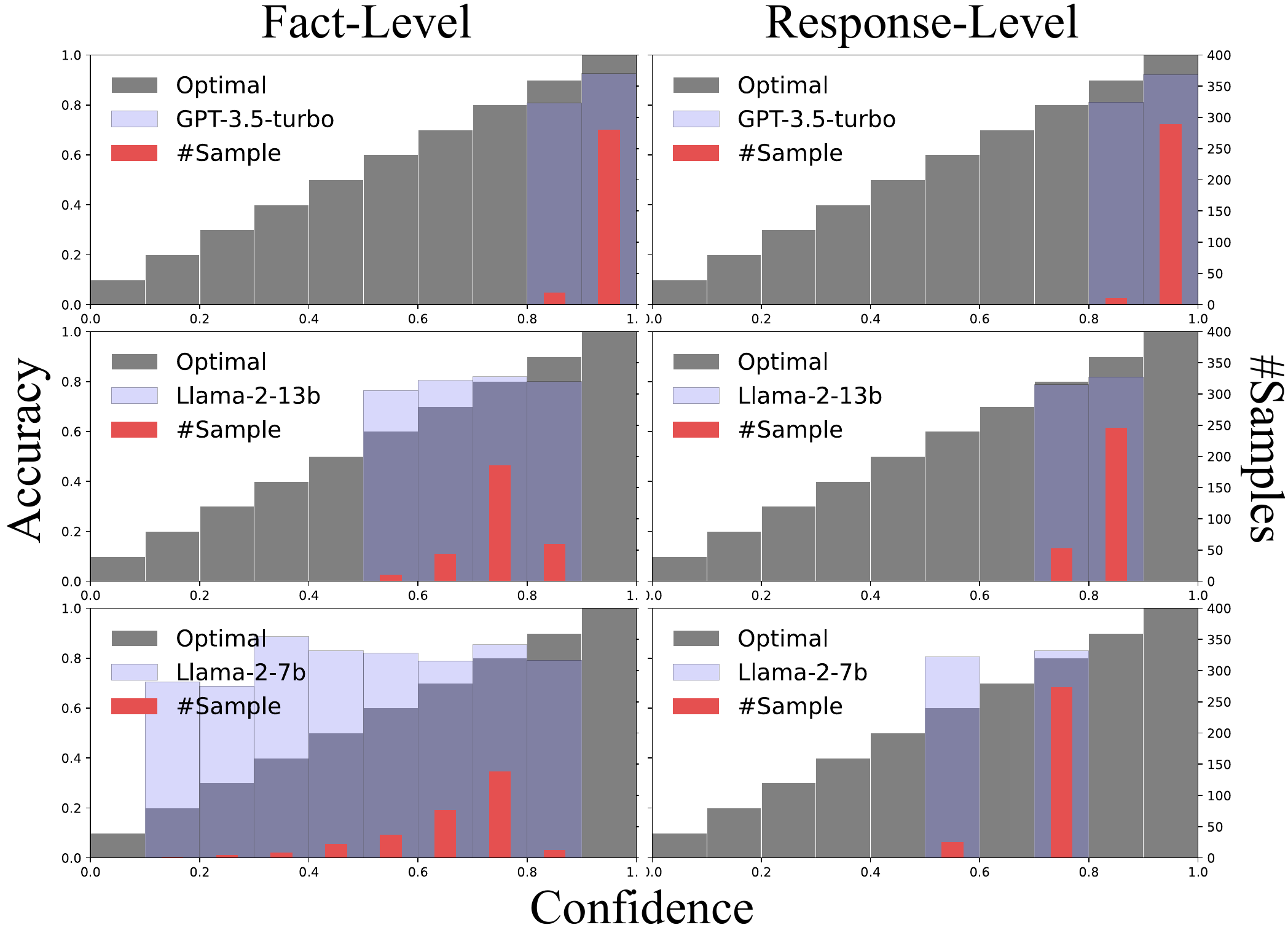}
    \vskip -0.5em
    \caption{Comparison of calibration measures between fact-level and response-level across models of three different scales: LLaMA-7B/13B, and GPT-3.5, indicating fact-level imposes a stricter standard (Observation 1).}
    \label{fig:ob-hist}
\end{figure}

\section{Key Observation}
\label{sec:observations}
This section discusses three important phenomena observed under our fact-level calibration.%
These findings not only demonstrate the superiority of our framework over traditional response-level calibration, but also inspire the development of a confidence-guided fact-level self-correction method based on these insights.

\subsection{Experiment Setup}
This section outlines the experimental setups, including the datasets, models, and evaluations.

\paragraph{Datasets} We utilize two long-form response datasets:
(1) LongFact~\cite{wei2024long}: a dataset contains prompts that evaluate model's factuality in open-end responses.
(2) ASQA~\cite{stelmakh2022asqa}: a dataset centers on ambiguous factoid questions with multiple reference answers.

\paragraph{Models} 
We use five popular models from different families and scales, including:
(1) LLaMA~\cite{DBLP:journals/corr/abs-2307-09288}: include LLaMA-7b-chat and LLaMA-13b-chat.
(2) Vicuna~\cite{vicuna2023}: include Vicuna-7b and Vicuna-13b.
(3) GPT~\cite{DBLP:conf/nips/BrownMRSKDNSSAA20}: GPT-3.5-turbo.

\paragraph{Correctness and Relevance Estimation} 
We use the Search-Augmented Factuality Evaluator (SAFE), a pipeline proposed by~\cite{wei2024long} that employs LLMs as agents to automatically evaluate the factuality of long-form responses. It utilizes a multi-step reasoning process that includes sending search queries to Google Search~\cite{hillis2012google} to verify the information provided. 

\paragraph{Confidence Estimation}
We utilize three popular confidence estimation methods from different categories~\cite{geng2024survey}: Verbalization~\cite{tian2023just}, isTrueLogit~\cite{DBLP:journals/corr/abs-2207-05221}, and Consistency~\cite{manakul-etal-2023-selfcheckgpt}.

\paragraph{Calibration Evaluation Metrics}
We use Expected Calibration Error (ECE)~\cite{guo2017calibration,Naeini2015ObtainingWC} at response-level, and F-ECE at the fact-level, as introduced in~\cref{calibration metric}.

\subsection{Results on Fact-Level Calibration}
\label{sec:observations}
This section highlights three key phenomena observed during our fact-level calibration. 
These findings not only underscore the advantages of our framework compared to traditional response-level's, but also inform the design of a confidence-guided fact-level self-correction approach.

\begin{figure*}[t]
    \centering
    \includegraphics[width=\linewidth]{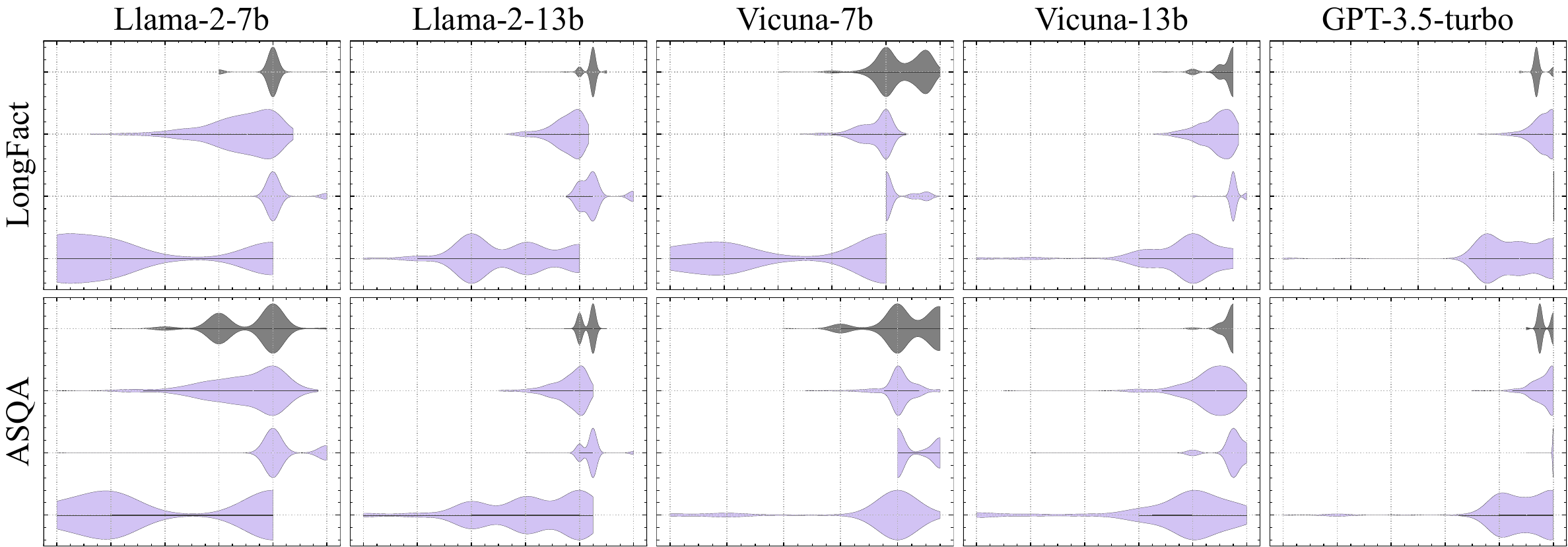}
    \vskip -0.5em
    \caption{Comparison of confidence distribution across different responses between fact-level and response-level, with the purple plots representing fact-level distribution under different statistical metrics and the gray plots showing the response-level distribution, highlighting the over-confidence issue at the response level, which stems from the dominance of the \textbf{implicit} high-confidence facts hidden within the response (Observation 2).}
    \label{fig:exp-violin}
\end{figure*}

\begin{figure*}[t]
    \centering
    \includegraphics[width=\linewidth]{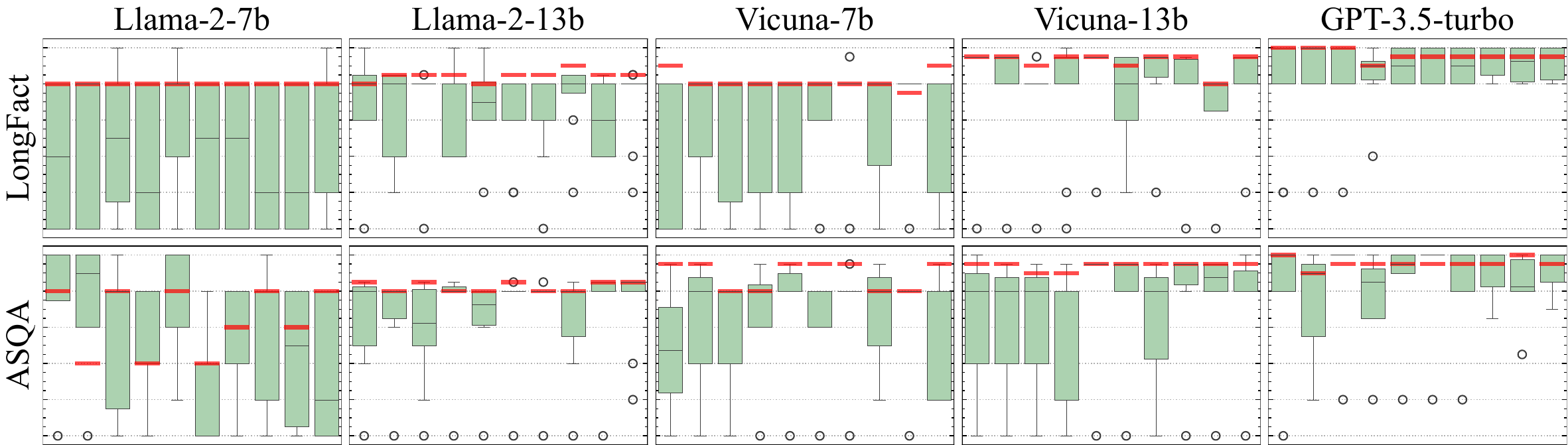}
    \vskip -0.5em
    \caption{Confidence distribution within individual responses at fact level, with red bar indicating the response-level confidence score, indicating high variance exists in fact-level confidence within a single response (Observation 3).}
    \label{fig:exp-box}
    \vspace{-4mm}
\end{figure*}

\paragraph{Observation 1: Fact-level calibration imposes a stricter standard.} 

We compare fact-level and response-level calibration following the protocol outlined in~\cite{guo2017calibration}, where reliability histograms and statistical metrics for is summarized.
For both fact and response level, we estimate confidence through all three confidence estimation method, where the correctness/relevance evaluation for facts are described in~\cref{sec:cal_frame}. 
The detailed prompt template is available in Appendix~\ref{app:prompts}. 
To generate the reliability histograms, we divide the model's predictions into ten bins based on confidence scores and calculate the average accuracy for each bin. Ideally, an optimally calibrated model would produce a diagonal pattern in the histogram.

As shown in~\cref{fig:ob-hist}, a comparison of the histograms on the left and right demonstrates that our fact-level framework can better distinguish calibration performance across models of varying scales. Notably, models such as Llama-2-7b, which appear well-calibrated under traditional response-level calibration, perform significantly worse under fact-level calibration. 
A similar trend is consistently observed in~\cref{tab:ece}, where, as highlighted in the table, the variance in fact-level calibration across five different models is nearly an order of magnitude higher than that of response-level calibration. 
This ability to reveal discrepancies arises from fact-level calibration's consideration of fine-grained correctness for each fact, as well as the relevance of each fact to the query. 
The alignment with the model's capabilities highlights both the superiority and validity of our method, indicating that fact-level calibration imposes a stricter standard.

\paragraph{Observation 2: Understanding over-confidence through fact-level analysis.}

We examine the distribution of sentence confidence for both fact-level and response-level calibrations. 
For the response-level, we directly plot its confidence scores distribution in a grey violin plot. 
For the fact-level, since confidence for a single response is represented as a vector rather than a scalar, we compute three statistical measures: the mean, maximum, and minimum of the vector. 
These measures are then visualized as three separate purple violin plots. 

The results, shown in~\cref{fig:exp-violin}, reveal two noteworthy phenomena:
(1) The confidence distribution at the response level is \underline{narrow}, concentrated around higher confidence values, while the distribution of the mean confidence in the fact-level calibration is \underline{wider} and indicates lower overall confidence.
(2) The response-level confidence distribution closely \underline{mirrors} the distribution of the maximum confidence values in the fact-level method.
Furthermore, a similar phenomenon is also corroborated in~\cref{tab:ece}, where the average confidence at the fact level across five different models is generally lower than that at the response level.

These observations suggest that response-level confidence is largely influenced by the fact with the highest implicit confidence, which can result in overconfidence. 
In contrast, fact-level framework breaks down the response into individual facts, assessing confidence for each one. This approach highlights less confident aspects of the response, helping to mitigate the issue of overconfidence.

\paragraph{Observation 3: High variance exists in fact-level confidence within a response.}
We examine the variance in fact-level confidence within each individual responses. 
Specifically, for each response, we extract its confidence vector and visualize its distribution using box plots. 
Due to space constraints, we have visualized 10 responses for each model across both datasets, as shown in~\cref{fig:exp-box}. 
Red bar represents the response-level confidence score.

Two key observations can be made:
(1) Fact-level confidence exhibits significant variance within individual responses, whereas response-level confidence tends to be concentrated at a higher level.
(2) Outlier facts typically have much lower confidence scores. The numerous white dots in the box plots indicate these outliers, which generally correspond to facts with substantially lower confidence, falling below the main distribution. 
This suggests that certain facts are generated by the model with considerably less confidence compared to others within the same response.

\section{Application: Confidence-Guided Fact-Level Self-Correction}

\begin{figure*}[t]
    \centering
    \includegraphics[width=\linewidth]{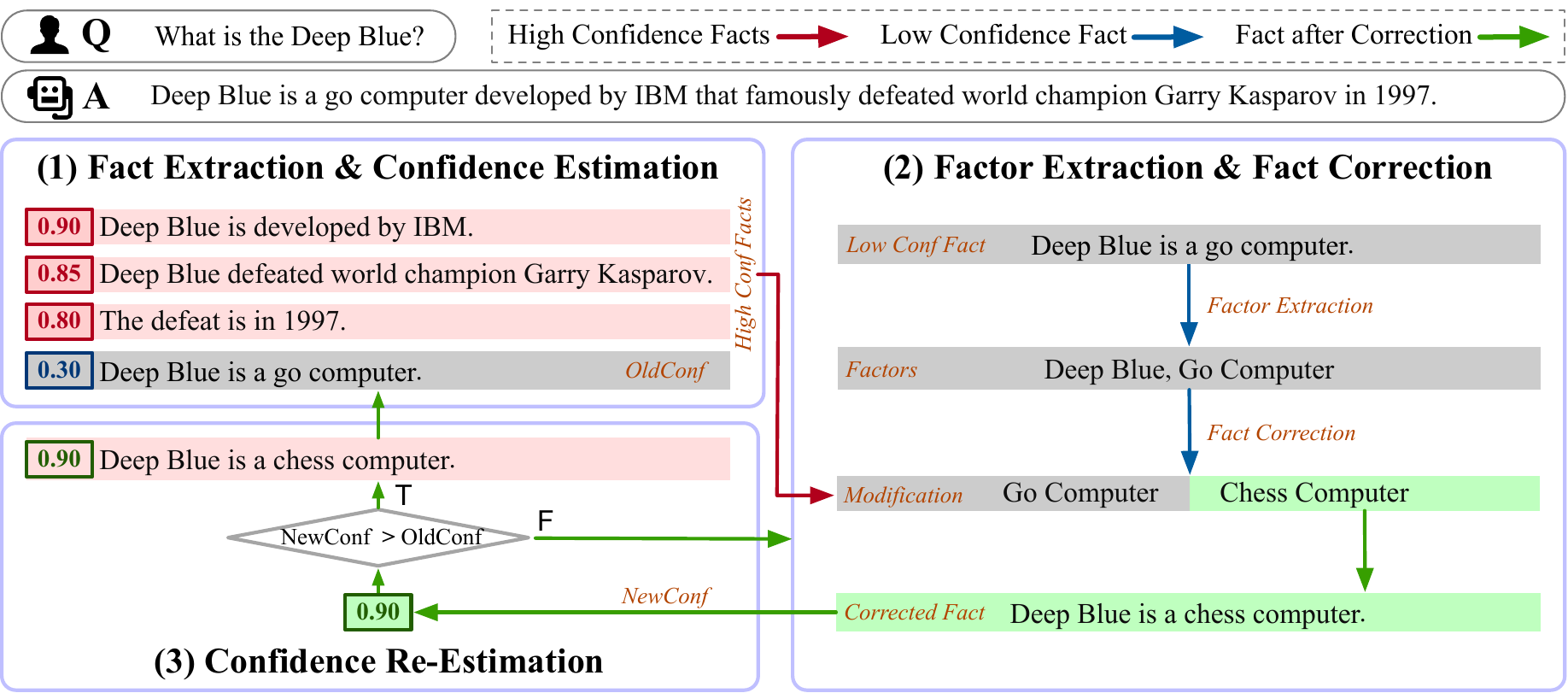}
    \vskip -0.5em
    \caption{An illustration of our confidence-guided fact-level self-correction framework (\texttt{ConFix}), consisting of three steps: (1) fact extraction and confidence estimation; (2) fact correction using high-confidence facts; and (3) confidence re-estimation. These three steps can be iterated until the desired level of enhancement is achieved.}
    \label{fig:self-corr}
\end{figure*}

In this section, we introduce the motivation and architecture of Confidence-Guided Fact-Level Self-Correction, dubbed \textbf{\texttt{ConFix}}. 
\texttt{ConFix} utilizes facts with high confidence as references to revise facts with low confidence, thereby enhancing the generation quality and mitigating hallucinations.
Importantly, \texttt{ConFix} operates in real-time during the generation process, eliminating the need for additional fine-tuning or retraining, which reduces computational costs and enhances its adaptability. Furthermore, it functions without relying on external knowledge sources, significantly broadening its applicability across diverse scenarios.

\subsection{Motivation}

The development of Confidence-Guided LLM Self-Correction is driven by the three observations discussed earlier. 
The rationale behind these observations supporting self-correction is as follows:
(1) Observations 1 and 2 demonstrate that, even under strict conditions, the fact-level framework reduces overconfidence and enhances the model’s calibration, aligning confidence more closely with factual accuracy. This improved calibration is crucial for enabling effective confidence-guided self-correction.
(2) Observation 3 reveals that high-confidence and low-confidence facts often coexist within the same response. 
Even when the overall confidence is relatively consistent, outliers tend to be low-confidence facts. 
This enables high-confidence facts to serve as some kind of references, providing the necessary information to refine and correct the low-confidence facts.

\subsection{Architecture}
The overall architecture of \texttt{ConFix} is illustrated in~\cref{fig:self-corr}, including three steps: (1) fact extraction \& confidence estimation, (2 )factor extraction \& fact correction, and (3) fact confidence re-estimation. 

\paragraph{S1: Fact Extraction \& Confidence Estimation}
Given a response $\mathbf{y}_i$, \texttt{ConFix} first conducts fact extraction and confidence estimation for each extracted fact, following the same process as described in ~\cref{sec:cal_frame}. 
After obtaining the facts $\{f_i^{j}\}_{j=1}^{M_i}$ for $\mathbf{y}_i$ and their corresponding confidence scores $\{\mathrm{conf}_i^{j}\}_{j=1}^{M_i}$, we then split the facts into two groups: high-confidence and low-confidence, based on a confidence threshold $\tau$.
The high-confidence group shown in~\cref{eq:high-conf} is used as a form of internal knowledge base, whose knowledge is leveraged to reinforce and augment facts within the low-confidence group, which is shown in~\cref{eq:low-conf},
\begin{align}
\mathit{f_{h}} = \{f_i^{j} \mid \mathrm{conf}_i^{j} \geq \tau\} \label{eq:high-conf} \\
\mathit{f_{l}} = \{f_i^{j} \mid \mathrm{conf}_i^{j} < \tau\}, \label{eq:low-conf}
\end{align}
where the threshold is defined as the mean confidence score across facts $\tau = \frac{1}{M_i} \sum_{j=1}^{M_i} \mathrm{conf}_i^{j}$.

\paragraph{S2: Factor Extraction \& Fact Correction}
To ensure that only the erroneous parts of the low-confidence facts are modified without changing the overall meaning, we restrict the modifiable parts. 
Specifically, we first parse the key factors through factor extraction. 
Let $\{\mathit{fa}_i^{j,k}\}_{k=1}^{K_i^j}$ represent the $K_i^j$ factors extracted from the target fact $f_i^{j} \in f_l$,
\begin{align}
\{\mathit{fa}_i^{j,k}\}_{k=1}^{K_i^j} = \mathcal{F}(\mathrm{LLM}(\cdot), p_f(f_i^j)),
\end{align}
where $p_f$ is the prompt, which includes:
(1) A clear task description.
(2) Several instances.
(3) The task containing the input sentence.
The model is expected to output its extracted factors. 

After extracting factors, we then perform fact correction, targeting only the extracted factors for modification. This process can be represented as,
\begin{align}
\hat{f}_i^{j} = \mathcal{R}(\mathrm{LLM}(\cdot), p_r(f_i^{j}, \{\mathit{fa}_i^{j,k}\}_{k=1}^{K_i^j}, f_h)
\end{align}
where $p_r$ is the prompt, which includes:
(1) A clear task description.
(2) Several instances.
(3) The task containing the input target fact, the extracted factors and the high-confidence reference facts.
The model is expected to output the modified target fact, noting that the model allows for returning ``NoError'' to make no modifications to the input.

\paragraph{S3: Fact Confidence Re-Estimation}
Finally, the modified facts undergo the confidence estimation process again to obtain new confidence scores:
\begin{align}
\hat{\mathrm{conf}}_i^j = \mathcal{C}\left(\mathrm{LLM}(\cdot), p_c(\hat{f}_i^j, \mathbf{x}_i, \mathbf{y}_i)\right),
\end{align}
where $\hat{\mathrm{conf}}_i^j$ represents the confidence score of the modified fact $ \hat{f}_i^j$.
Finally, if $ \hat{\mathrm{conf}}_i^j > \mathrm{conf}_i^j $, the modification is deemed successful and is accepted. Otherwise, \texttt{ConFix} will repeat the process of factor extraction, fact correction, and confidence re-estimation. 
This iterative process continues until either a satisfactory confidence score is achieved or a predetermined maximum number of iterations $ N $ is reached, where the model return ``NoError'' and make no modifications to the input.

\subsection{Experiment on Self-Correction}

\paragraph{Experiment Setup}
We adopt a two-fold  evaluation. First, we assess error detection performance using standard metrics such as Accuracy, Precision, and Recall~\cite{powers2020evaluation}. Subsequently, for evaluating self-correction, we leverage GPT-4 in a zero-shot pairwise evaluation framework (see prompts in Appendix~\ref{app:prompts}), measuring improvement ratio, consistency ratio, and regression ratio to capture the performance of correction.

\paragraph{Error Detection}
Table~\ref{tab:detect} shows the error detection results of our proposed method across five different base models. 
As seen from the table, larger models, such as GPT and the 13B model, outperform smaller models like the 7B model in terms of Accuracy and Precision. 
However, all models show lower performance in Recall, suggesting that a significant number of erroneous facts go undetected. 
This indicates a persistent overconfidence issue, with models often incorrectly considering false information as correct.

\paragraph{Error Correction}
Table~\ref{tab:self_correction} presents the error correction results of our method across five different base models. 
The results show that, among the three possible outcomes: Improved, Same, and Regressed, our method consistently achieves the highest proportion of improve for almost all models, with the exception of LLaMA-2-7B. 

The performance drop observed in LLaMA-2-7B is interesting, as it aligns with findings from previous experiments:
\texttt{ConFix} depends on the model’s calibration capabilities, functioning effectively when confidence is well-aligned with correctness.
However, LLaMA-2-7B is inadequately calibrated, as indicated by its higher ECE/F-ECE scores in~\cref{tab:ece}. 
Therefore, the self-correction performance drop in LLaMA-2-7B does not undermine our \texttt{ConFix}, but rather highlights the importance of adequate calibration. \underline{Under the right conditions}, our method can significantly improve model performance without the need for external knowledge, and these conditions are easily met, as calibration can be enhanced using various existing techniques~\cite{liu2023lightweight,geng2024survey}.

\begin{table}[h]
\centering
\caption{Acc., Precision and Recall of Error-Detection.}
\vskip -0.5em
\label{tab:model_comparison}
\begin{adjustbox}{width=\linewidth}
\setlength{\tabcolsep}{1mm}
\begin{tabular}{lccc}
\toprule
\textbf{Base Model} & \textbf{Accuracy} (\%)  & \textbf{Precision} (\%)  & \textbf{Recall} (\%)  \\ 
\midrule
GPT-3.5-turbo & 83.29 & 87.89 & 13.71 \\ 
Vicuna-7b & 60.06 & 99.90 & 0.15  \\ 
Vicuna-13b & 74.81 & 77.68 & 8.46 \\ 
Llama-2-7b & 64.26 & 67.86 & 13.35\\ 
Llama-2-13b & 70.45 & 77.45 & 30.62\\ 
\bottomrule
\end{tabular}
\label{tab:detect}
\end{adjustbox}
\end{table}

\vskip -0.9em
\begin{table}[h]
\centering
\caption{GPT-4 Evaluation of Self-Correction.}
\vskip -0.5em
\label{tab:model_comparison}
\begin{adjustbox}{width=\linewidth}
\setlength{\tabcolsep}{1mm}
\begin{tabular}{lcccc}
\toprule
\textbf{Base Model} & \textbf{Improved} (\%) & \textbf{Same} (\%)& \textbf{Regressed} (\%) & \textbf{\#Revised}\\ 
\midrule
GPT-3.5-turbo & 46.30 & 24.07 & 29.63 & 108 \\ 

Vicuna-7b & 50.00  & 50.00 & 0.00 & 2 \\ 
Vicuna-13b & 49.40 & 28.92 & 21.69 & 83 \\
Llama-2-7b & 6.76 & 12.56 & 80.68 & 207 \\ 
Llama-2-13b & 53.35 & 19.59 & 27.07 & 418\\ 
\bottomrule
\end{tabular}
\label{tab:self_correction}
\end{adjustbox}
\vspace{-2mm}
\end{table}

\section{Conclusion}

This paper presents a novel fact-level calibration and self-correction framework designed to address hallucination issues in long-form responses generated by large language models. 
Traditional response-level methods fall short when handling complex outputs containing multiple facts. 
Our approach evaluates each fact's correctness and relevance individually, both externally and internally, enabling fine-grained confidence assessments. 
This framework imposes a stricter standard than response-level methods, reduces overconfidence, and highlights significant confidence variance across facts within a response. 
By leveraging high-confidence facts for in-context learning, our self-correction method \texttt{ConFix} can effectively mitigate hallucinations without the need for external knowledge, as demonstrated across various models.

\clearpage
\section{Limitations and Broader Impacts}
In this work, we propose a fact-level calibration framework and, based on this framework, introduce a confidence-guided fact-level self-correction method. However, for this self-correction method to be effective, the model itself must possess a certain level of calibration ability.
In our paper, we discuss how our calibration framework can alleviate over-confidence. In future work, we will further explore ways to enhance calibration ability within the calibration framework, paving the way for more effective confidence-guided self-correction.
\bibliography{main}

\begin{thebibliography}{40}
\providecommand{\natexlab}[1]{#1}

\bibitem[{Bang et~al.(2023)Bang, Cahyawijaya, Lee, Dai, Su, Wilie, Lovenia, Ji, Yu, Chung et~al.}]{bang2023multitask}
Yejin Bang, Samuel Cahyawijaya, Nayeon Lee, Wenliang Dai, Dan Su, Bryan Wilie, Holy Lovenia, Ziwei Ji, Tiezheng Yu, Willy Chung, et~al. 2023.
\newblock A multitask, multilingual, multimodal evaluation of chatgpt on reasoning, hallucination, and interactivity.
\newblock \emph{arXiv preprint arXiv:2302.04023}.

\bibitem[{Brown et~al.(2020)Brown, Mann, Ryder, Subbiah, Kaplan, Dhariwal, Neelakantan, Shyam, Sastry, Askell, Agarwal, Herbert{-}Voss, Krueger, Henighan, Child, Ramesh, Ziegler, Wu, Winter, Hesse, Chen, Sigler, Litwin, Gray, Chess, Clark, Berner, McCandlish, Radford, Sutskever, and Amodei}]{DBLP:conf/nips/BrownMRSKDNSSAA20}
Tom~B. Brown, Benjamin Mann, Nick Ryder, Melanie Subbiah, Jared Kaplan, Prafulla Dhariwal, Arvind Neelakantan, Pranav Shyam, Girish Sastry, Amanda Askell, Sandhini Agarwal, Ariel Herbert{-}Voss, Gretchen Krueger, Tom Henighan, Rewon Child, Aditya Ramesh, Daniel~M. Ziegler, Jeffrey Wu, Clemens Winter, Christopher Hesse, Mark Chen, Eric Sigler, Mateusz Litwin, Scott Gray, Benjamin Chess, Jack Clark, Christopher Berner, Sam McCandlish, Alec Radford, Ilya Sutskever, and Dario Amodei. 2020.
\newblock \href {https://proceedings.neurips.cc/paper/2020/hash/1457c0d6bfcb4967418bfb8ac142f64a-Abstract.html} {Language models are few-shot learners}.
\newblock In \emph{Advances in Neural Information Processing Systems 33: Annual Conference on Neural Information Processing Systems 2020, NeurIPS 2020, December 6-12, 2020, virtual}.

\bibitem[{Cheng et~al.(2023)Cheng, Gan, Yang, Wang, Wang, Boyd-Graber, and Wang}]{cheng2023prompting}
Silei Cheng, Zhe Gan, Zhengyuan Yang, Shuohang Wang, Jianfeng Wang, Jordan Boyd-Graber, and Lijuan Wang. 2023.
\newblock \href {https://www.microsoft.com/en-us/research/publication/prompting-gpt-3-to-be-reliable/} {Prompting gpt-3 to be reliable}.
\newblock In \emph{International Conference on Learning Representations (ICLR 23)}.

\bibitem[{Chiang et~al.(2023)Chiang, Li, Lin, Sheng, Wu, Zhang, Zheng, Zhuang, Zhuang, Gonzalez, Stoica, and Xing}]{vicuna2023}
Wei-Lin Chiang, Zhuohan Li, Zi~Lin, Ying Sheng, Zhanghao Wu, Hao Zhang, Lianmin Zheng, Siyuan Zhuang, Yonghao Zhuang, Joseph~E. Gonzalez, Ion Stoica, and Eric~P. Xing. 2023.
\newblock \href {https://lmsys.org/blog/2023-03-30-vicuna/} {Vicuna: An open-source chatbot impressing gpt-4 with 90\%* chatgpt quality}.

\bibitem[{Dan and Roth(2021)}]{dan2021effects}
Soham Dan and Dan Roth. 2021.
\newblock On the effects of transformer size on in-and out-of-domain calibration.
\newblock In \emph{Findings of the Association for Computational Linguistics: EMNLP 2021}, pages 2096--2101.

\bibitem[{Desai and Durrett(2020)}]{desai2020calibration}
Shrey Desai and Greg Durrett. 2020.
\newblock Calibration of pre-trained transformers.
\newblock \emph{arXiv preprint arXiv:2003.07892}.

\bibitem[{Dubois et~al.(2024)Dubois, Galambosi, Liang, and Hashimoto}]{dubois2024length}
Yann Dubois, Bal{\'a}zs Galambosi, Percy Liang, and Tatsunori~B Hashimoto. 2024.
\newblock Length-controlled alpacaeval: A simple way to debias automatic evaluators.
\newblock \emph{arXiv preprint arXiv:2404.04475}.

\bibitem[{Dubois et~al.(2023)Dubois, Li, Taori, Zhang, Gulrajani, Ba, Guestrin, Liang, and Hashimoto}]{dubois2023alpacafarm}
Yann Dubois, Xuechen Li, Rohan Taori, Tianyi Zhang, Ishaan Gulrajani, Jimmy Ba, Carlos Guestrin, Percy Liang, and Tatsunori~B. Hashimoto. 2023.
\newblock \href {https://arxiv.org/abs/2305.14387} {Alpacafarm: A simulation framework for methods that learn from human feedback}.
\newblock \emph{Preprint}, arXiv:2305.14387.

\bibitem[{Geng et~al.(2024)Geng, Cai, Wang, Koeppl, Nakov, and Gurevych}]{geng2024survey}
Jiahui Geng, Fengyu Cai, Yuxia Wang, Heinz Koeppl, Preslav Nakov, and Iryna Gurevych. 2024.
\newblock A survey of confidence estimation and calibration in large language models.
\newblock In \emph{Proceedings of the 2024 Conference of the North American Chapter of the Association for Computational Linguistics: Human Language Technologies (Volume 1: Long Papers)}, pages 6577--6595.

\bibitem[{Golovneva et~al.(2022)Golovneva, Chen, Poff, Corredor, Zettlemoyer, Fazel-Zarandi, and Celikyilmaz}]{golovneva2022roscoe}
Olga Golovneva, Moya~Peng Chen, Spencer Poff, Martin Corredor, Luke Zettlemoyer, Maryam Fazel-Zarandi, and Asli Celikyilmaz. 2022.
\newblock Roscoe: A suite of metrics for scoring step-by-step reasoning.
\newblock In \emph{The Eleventh International Conference on Learning Representations}.

\bibitem[{Guo et~al.(2017{\natexlab{a}})Guo, Pleiss, Sun, and Weinberger}]{guo2017calibration}
Chuan Guo, Geoff Pleiss, Yu~Sun, and Kilian~Q Weinberger. 2017{\natexlab{a}}.
\newblock On calibration of modern neural networks.
\newblock In \emph{International conference on machine learning}, pages 1321--1330. PMLR.

\bibitem[{Guo et~al.(2017{\natexlab{b}})Guo, Pleiss, Sun, and Weinberger}]{DBLP:conf/icml/GuoPSW17}
Chuan Guo, Geoff Pleiss, Yu~Sun, and Kilian~Q. Weinberger. 2017{\natexlab{b}}.
\newblock \href {http://proceedings.mlr.press/v70/guo17a.html} {On calibration of modern neural networks}.
\newblock In \emph{Proceedings of the 34th International Conference on Machine Learning, {ICML} 2017, Sydney, NSW, Australia, 6-11 August 2017}, volume~70 of \emph{Proceedings of Machine Learning Research}, pages 1321--1330. {PMLR}.

\bibitem[{Hillis et~al.(2012)Hillis, Petit, and Jarrett}]{hillis2012google}
Ken Hillis, Michael Petit, and Kylie Jarrett. 2012.
\newblock \emph{Google and the Culture of Search}.
\newblock Routledge.

\bibitem[{Hu et~al.(2023)Hu, Zhang, Zhao, Huang, and Wu}]{hu2023uncertainty}
Mengting Hu, Zhen Zhang, Shiwan Zhao, Minlie Huang, and Bingzhe Wu. 2023.
\newblock Uncertainty in natural language processing: Sources, quantification, and applications.
\newblock \emph{arXiv preprint arXiv:2306.04459}.

\bibitem[{Huang et~al.(2024)Huang, Liu, Thirukovalluru, Cohan, and Dhingra}]{huang2024calibrating}
Yukun Huang, Yixin Liu, Raghuveer Thirukovalluru, Arman Cohan, and Bhuwan Dhingra. 2024.
\newblock Calibrating long-form generations from large language models.
\newblock \emph{arXiv preprint arXiv:2402.06544}.

\bibitem[{Kadavath et~al.(2022{\natexlab{a}})Kadavath, Conerly, Askell, Henighan, Drain, Perez, Schiefer, Hatfield{-}Dodds, DasSarma, Tran{-}Johnson, Johnston, Showk, Jones, Elhage, Hume, Chen, Bai, Bowman, Fort, Ganguli, Hernandez, Jacobson, Kernion, Kravec, Lovitt, Ndousse, Olsson, Ringer, Amodei, Brown, Clark, Joseph, Mann, McCandlish, Olah, and Kaplan}]{DBLP:journals/corr/abs-2207-05221}
Saurav Kadavath, Tom Conerly, Amanda Askell, Tom Henighan, Dawn Drain, Ethan Perez, Nicholas Schiefer, Zac Hatfield{-}Dodds, Nova DasSarma, Eli Tran{-}Johnson, Scott Johnston, Sheer~El Showk, Andy Jones, Nelson Elhage, Tristan Hume, Anna Chen, Yuntao Bai, Sam Bowman, Stanislav Fort, Deep Ganguli, Danny Hernandez, Josh Jacobson, Jackson Kernion, Shauna Kravec, Liane Lovitt, Kamal Ndousse, Catherine Olsson, Sam Ringer, Dario Amodei, Tom Brown, Jack Clark, Nicholas Joseph, Ben Mann, Sam McCandlish, Chris Olah, and Jared Kaplan. 2022{\natexlab{a}}.
\newblock \href {https://doi.org/10.48550/arXiv.2207.05221} {Language models (mostly) know what they know}.
\newblock \emph{CoRR}, abs/2207.05221.

\bibitem[{Kadavath et~al.(2022{\natexlab{b}})Kadavath, Conerly, Askell, Henighan, Drain, Perez, Schiefer, Hatfield-Dodds, DasSarma, Tran-Johnson et~al.}]{kadavath2022language}
Saurav Kadavath, Tom Conerly, Amanda Askell, Tom Henighan, Dawn Drain, Ethan Perez, Nicholas Schiefer, Zac Hatfield-Dodds, Nova DasSarma, Eli Tran-Johnson, et~al. 2022{\natexlab{b}}.
\newblock Language models (mostly) know what they know.
\newblock \emph{arXiv preprint arXiv:2207.05221}.

\bibitem[{Kuhn et~al.(2023)Kuhn, Gal, and Farquhar}]{DBLP:conf/iclr/KuhnGF23}
Lorenz Kuhn, Yarin Gal, and Sebastian Farquhar. 2023.
\newblock \href {https://openreview.net/pdf?id=VD-AYtP0dve} {Semantic uncertainty: Linguistic invariances for uncertainty estimation in natural language generation}.
\newblock In \emph{The Eleventh International Conference on Learning Representations, {ICLR} 2023, Kigali, Rwanda, May 1-5, 2023}. OpenReview.net.

\bibitem[{Li et~al.(2023{\natexlab{a}})Li, Cheng, Zhao, Nie, and Wen}]{li2023halueval}
Junyi Li, Xiaoxue Cheng, Wayne~Xin Zhao, Jian-Yun Nie, and Ji-Rong Wen. 2023{\natexlab{a}}.
\newblock Halueval: A large-scale hallucination evaluation benchmark for large language models.
\newblock In \emph{Proceedings of the 2023 Conference on Empirical Methods in Natural Language Processing}, pages 6449--6464.

\bibitem[{Li et~al.(2024)Li, Wang, Feng, Zhu, Wang, and Chua}]{li2024think}
Moxin Li, Wenjie Wang, Fuli Feng, Fengbin Zhu, Qifan Wang, and Tat-Seng Chua. 2024.
\newblock Think twice before assure: Confidence estimation for large language models through reflection on multiple answers.
\newblock \emph{arXiv preprint arXiv:2403.09972}.

\bibitem[{Li et~al.(2023{\natexlab{b}})Li, Zhang, Dubois, Taori, Gulrajani, Guestrin, Liang, and Hashimoto}]{alpaca_eval}
Xuechen Li, Tianyi Zhang, Yann Dubois, Rohan Taori, Ishaan Gulrajani, Carlos Guestrin, Percy Liang, and Tatsunori~B. Hashimoto. 2023{\natexlab{b}}.
\newblock Alpacaeval: An automatic evaluator of instruction-following models.
\newblock \url{https://github.com/tatsu-lab/alpaca_eval}.

\bibitem[{Lin et~al.(2021)Lin, Hilton, and Evans}]{lin2021truthfulqa}
Stephanie Lin, Jacob Hilton, and Owain Evans. 2021.
\newblock Truthfulqa: Measuring how models mimic human falsehoods.
\newblock \emph{arXiv preprint arXiv:2109.07958}.

\bibitem[{Liu et~al.(2023)Liu, Khalifa, and Wang}]{liu2023lightweight}
Xin Liu, Muhammad Khalifa, and Lu~Wang. 2023.
\newblock Lightweight language model calibration for open-ended question answering with varied answer lengths.
\newblock In \emph{The Twelfth International Conference on Learning Representations}.

\bibitem[{Manakul et~al.(2023)Manakul, Liusie, and Gales}]{manakul-etal-2023-selfcheckgpt}
Potsawee Manakul, Adian Liusie, and Mark Gales. 2023.
\newblock \href {https://doi.org/10.18653/v1/2023.emnlp-main.557} {{S}elf{C}heck{GPT}: Zero-resource black-box hallucination detection for generative large language models}.
\newblock In \emph{Proceedings of the 2023 Conference on Empirical Methods in Natural Language Processing}, pages 9004--9017, Singapore. Association for Computational Linguistics.

\bibitem[{Min et~al.(2023)Min, Krishna, Lyu, Lewis, Yih, Koh, Iyyer, Zettlemoyer, and Hajishirzi}]{DBLP:journals/corr/abs-2305-14251}
Sewon Min, Kalpesh Krishna, Xinxi Lyu, Mike Lewis, Wen{-}tau Yih, Pang~Wei Koh, Mohit Iyyer, Luke Zettlemoyer, and Hannaneh Hajishirzi. 2023.
\newblock \href {https://doi.org/10.48550/arXiv.2305.14251} {Factscore: Fine-grained atomic evaluation of factual precision in long form text generation}.
\newblock \emph{CoRR}, abs/2305.14251.

\bibitem[{M{\"u}ndler et~al.(2023)M{\"u}ndler, He, Jenko, and Vechev}]{mundler2023self}
Niels M{\"u}ndler, Jingxuan He, Slobodan Jenko, and Martin Vechev. 2023.
\newblock Self-contradictory hallucinations of large language models: Evaluation, detection and mitigation.
\newblock \emph{arXiv preprint arXiv:2305.15852}.

\bibitem[{Naeini et~al.(2015)Naeini, Cooper, and Hauskrecht}]{Naeini2015ObtainingWC}
Mahdi~Pakdaman Naeini, Gregory~F. Cooper, and Milos Hauskrecht. 2015.
\newblock \href {https://api.semanticscholar.org/CorpusID:6292807} {Obtaining well calibrated probabilities using bayesian binning}.
\newblock \emph{Proceedings of the ... AAAI Conference on Artificial Intelligence. AAAI Conference on Artificial Intelligence}, 2015:2901--2907.

\bibitem[{Nguyen and O'Connor(2015)}]{DBLP:conf/emnlp/NguyenO15}
Khanh Nguyen and Brendan O'Connor. 2015.
\newblock \href {https://doi.org/10.18653/v1/d15-1182} {Posterior calibration and exploratory analysis for natural language processing models}.
\newblock In \emph{Proceedings of the 2015 Conference on Empirical Methods in Natural Language Processing, {EMNLP} 2015, Lisbon, Portugal, September 17-21, 2015}, pages 1587--1598. The Association for Computational Linguistics.

\bibitem[{OpenAI(2023)}]{DBLP:journals/corr/abs-2303-08774}
OpenAI. 2023.
\newblock \href {https://doi.org/10.48550/arXiv.2303.08774} {{GPT-4} technical report}.
\newblock \emph{CoRR}, abs/2303.08774.

\bibitem[{Powers(2020)}]{powers2020evaluation}
David~MW Powers. 2020.
\newblock Evaluation: from precision, recall and f-measure to roc, informedness, markedness and correlation.
\newblock \emph{arXiv preprint arXiv:2010.16061}.

\bibitem[{Stelmakh et~al.(2022)Stelmakh, Luan, Dhingra, and Chang}]{stelmakh2022asqa}
Ivan Stelmakh, Yi~Luan, Bhuwan Dhingra, and Ming-Wei Chang. 2022.
\newblock Asqa: Factoid questions meet long-form answers.
\newblock \emph{arXiv preprint arXiv:2204.06092}.

\bibitem[{Tian et~al.(2023)Tian, Mitchell, Zhou, Sharma, Rafailov, Yao, Finn, and Manning}]{tian2023just}
Katherine Tian, Eric Mitchell, Allan Zhou, Archit Sharma, Rafael Rafailov, Huaxiu Yao, Chelsea Finn, and Christopher~D Manning. 2023.
\newblock Just ask for calibration: Strategies for eliciting calibrated confidence scores from language models fine-tuned with human feedback.
\newblock \emph{arXiv preprint arXiv:2305.14975}.

\bibitem[{Touvron et~al.(2023)Touvron, Martin, Stone, Albert, Almahairi, Babaei, Bashlykov, Batra, Bhargava, Bhosale, Bikel, Blecher, Canton{-}Ferrer, Chen, Cucurull, Esiobu, Fernandes, Fu, Fu, Fuller, Gao, Goswami, Goyal, Hartshorn, Hosseini, Hou, Inan, Kardas, Kerkez, Khabsa, Kloumann, Korenev, Koura, Lachaux, Lavril, Lee, Liskovich, Lu, Mao, Martinet, Mihaylov, Mishra, Molybog, Nie, Poulton, Reizenstein, Rungta, Saladi, Schelten, Silva, Smith, Subramanian, Tan, Tang, Taylor, Williams, Kuan, Xu, Yan, Zarov, Zhang, Fan, Kambadur, Narang, Rodriguez, Stojnic, Edunov, and Scialom}]{DBLP:journals/corr/abs-2307-09288}
Hugo Touvron, Louis Martin, Kevin Stone, Peter Albert, Amjad Almahairi, Yasmine Babaei, Nikolay Bashlykov, Soumya Batra, Prajjwal Bhargava, Shruti Bhosale, Dan Bikel, Lukas Blecher, Cristian Canton{-}Ferrer, Moya Chen, Guillem Cucurull, David Esiobu, Jude Fernandes, Jeremy Fu, Wenyin Fu, Brian Fuller, Cynthia Gao, Vedanuj Goswami, Naman Goyal, Anthony Hartshorn, Saghar Hosseini, Rui Hou, Hakan Inan, Marcin Kardas, Viktor Kerkez, Madian Khabsa, Isabel Kloumann, Artem Korenev, Punit~Singh Koura, Marie{-}Anne Lachaux, Thibaut Lavril, Jenya Lee, Diana Liskovich, Yinghai Lu, Yuning Mao, Xavier Martinet, Todor Mihaylov, Pushkar Mishra, Igor Molybog, Yixin Nie, Andrew Poulton, Jeremy Reizenstein, Rashi Rungta, Kalyan Saladi, Alan Schelten, Ruan Silva, Eric~Michael Smith, Ranjan Subramanian, Xiaoqing~Ellen Tan, Binh Tang, Ross Taylor, Adina Williams, Jian~Xiang Kuan, Puxin Xu, Zheng Yan, Iliyan Zarov, Yuchen Zhang, Angela Fan, Melanie Kambadur, Sharan Narang, Aur{\'{e}}lien Rodriguez, Robert Stojnic, Sergey Edunov,
  and Thomas Scialom. 2023.
\newblock \href {https://doi.org/10.48550/arXiv.2307.09288} {Llama 2: Open foundation and fine-tuned chat models}.
\newblock \emph{CoRR}, abs/2307.09288.

\bibitem[{Wang et~al.(2024)Wang, Song, Peng, Tian, Jin, Mi, Su, and Yu}]{wang2024fine}
Ante Wang, Linfeng Song, Baolin Peng, Ye~Tian, Lifeng Jin, Haitao Mi, Jinsong Su, and Dong Yu. 2024.
\newblock Fine-grained self-endorsement improves factuality and reasoning.
\newblock \emph{arXiv preprint arXiv:2402.15631}.

\bibitem[{Wang et~al.(2023)Wang, Wei, Schuurmans, Le, Chi, Narang, Chowdhery, and Zhou}]{DBLP:conf/iclr/0002WSLCNCZ23}
Xuezhi Wang, Jason Wei, Dale Schuurmans, Quoc~V. Le, Ed~H. Chi, Sharan Narang, Aakanksha Chowdhery, and Denny Zhou. 2023.
\newblock \href {https://openreview.net/pdf?id=1PL1NIMMrw} {Self-consistency improves chain of thought reasoning in language models}.
\newblock In \emph{The Eleventh International Conference on Learning Representations, {ICLR} 2023, Kigali, Rwanda, May 1-5, 2023}. OpenReview.net.

\bibitem[{Wei et~al.(2024)Wei, Yang, Song, Lu, Hu, Tran, Peng, Liu, Huang, Du et~al.}]{wei2024long}
Jerry Wei, Chengrun Yang, Xinying Song, Yifeng Lu, Nathan Hu, Dustin Tran, Daiyi Peng, Ruibo Liu, Da~Huang, Cosmo Du, et~al. 2024.
\newblock Long-form factuality in large language models.
\newblock \emph{arXiv preprint arXiv:2403.18802}.

\bibitem[{Xiao et~al.(2024{\natexlab{a}})Xiao, Li, Yuan, Zhu, Cui, and Honavar}]{xiao2024leverage}
Teng Xiao, Mingxiao Li, Yige Yuan, Huaisheng Zhu, Chao Cui, and Vasant~G Honavar. 2024{\natexlab{a}}.
\newblock How to leverage demonstration data in alignment for large language model? a self-imitation learning perspective.
\newblock \emph{arXiv preprint arXiv:2410.10093}.

\bibitem[{Xiao et~al.(2024{\natexlab{b}})Xiao, Yuan, Zhu, Li, and Honavar}]{xiao2024cal}
Teng Xiao, Yige Yuan, Huaisheng Zhu, Mingxiao Li, and Vasant~G Honavar. 2024{\natexlab{b}}.
\newblock Cal-dpo: Calibrated direct preference optimization for language model alignment.
\newblock In \emph{The Thirty-eighth Annual Conference on Neural Information Processing Systems}.

\bibitem[{Zhang et~al.(2024)Zhang, Liu, Basaldella, and Collier}]{zhang2024luq}
Caiqi Zhang, Fangyu Liu, Marco Basaldella, and Nigel Collier. 2024.
\newblock Luq: Long-text uncertainty quantification for llms.
\newblock \emph{arXiv preprint arXiv:2403.20279}.

\bibitem[{Zhang et~al.(2023)Zhang, Li, Cui, Cai, Liu, Fu, Huang, Zhao, Zhang, Chen et~al.}]{zhang2023siren}
Yue Zhang, Yafu Li, Leyang Cui, Deng Cai, Lemao Liu, Tingchen Fu, Xinting Huang, Enbo Zhao, Yu~Zhang, Yulong Chen, et~al. 2023.
\newblock Siren's song in the ai ocean: A survey on hallucination in large language models.
\newblock \emph{arXiv preprint arXiv:2309.01219}.

\end{thebibliography}

\clearpage
\appendix

\section{Datasets}
\paragraph{LongFact}~\cite{wei2024long}: A dataset consisting of prompts designed to assess a model's factuality in long-form responses created by GPT-4. This dataset includes a diverse range of topics and ensures that the prompts require detailed and nuanced answers, making it a robust benchmark for evaluating the factual accuracy of language models in generating extended text. The dataset is particularly valuable for testing the capabilities of models in maintaining factual consistency over longer passages, which is crucial for applications such as content creation, summarization, and complex question answering.

\paragraph{ASQA}~\cite{stelmakh2022asqa}: A dataset designed for long-form question answering that uniquely centers on ambiguous factoid questions. ASQA provides a challenging testbed for models as it includes questions that can have multiple valid answers depending on the interpretation of the ambiguity. This dataset emphasizes the need for models to not only retrieve accurate information but also to handle the inherent uncertainty and provide comprehensive responses. ASQA is instrumental in pushing the boundaries of model performance in scenarios where clarity and precision are essential, such as in education and interactive AI systems.

\section{Models}
\paragraph{LLaMA-7b-chat \& LLaMA-13b-chat}~\cite{DBLP:journals/corr/abs-2307-09288}: These models are part of the LLaMA family, known for their strong performance in various natural language processing tasks. The ``chat'' versions are particularly fine-tuned for conversational contexts, making them suitable for generating coherent and contextually appropriate responses in dialogue settings. LLaMA models are designed to balance performance and computational efficiency, making it a popular choice for research and application in interactive AI systems.

\paragraph{Vicuna-7b and Vicuna-13b}~\cite{vicuna2023}
Vicuna is an open-source chatbot trained by fine-tuning LLaMA on user-shared conversations collected from ShareGPT.
Preliminary evaluation using GPT-4 as a judge shows Vicuna-13B achieves more than 90\% quality of ChatGPT and Bard while outperforming other models like LLaMA and Stanford Alpaca~\cite{alpaca_eval,dubois2024length,dubois2023alpacafarm} in more than 90\% of cases.

\paragraph{ GPT-3.5-turbo}~\cite{DBLP:conf/nips/BrownMRSKDNSSAA20}
This model is part of OpenAI's well-known GPT series. GPT-3.5-turbo is an enhanced version of GPT-3, offering improved performance and efficiency. It is designed to handle a wide range of language tasks, from text generation to comprehension and translation. The "turbo" variant is optimized for faster inference and lower latency, making it ideal for real-time applications where quick response times are crucial. GPT-3.5-turbo is widely used in both research and industry due to its versatility and high-quality output.

\section{Prompts}
\label{app:prompts}

\subsection{Prompt for Fact-Level Confidence Estimation}
The specific prompts used for fact-level confidence estimation are detailed below.
\begin{table}[h!]
    \small
    {\ttfamily
    \begin{tabularx}{\linewidth}{X}
    \toprule

    Instructions:\\
    1. The following STATEMENT has been extracted from the broader context of the given RESPONSE to the given QUESTION.\\
    2. Indicate how confident you are in the accuracy of the STATEMENT when answering the QUESTION, based on your knowledge.\\
    3. The confidence evaluation should be a value between 0 and 1 (with two decimal places retained), based on the following scoring criterion: \\
    \{Criterion\}
    4. Your task is to do this for the STATEMENT, RESPONSE and QUESTION under "Your Task". \\
    Some examples have been provided for you to learn how to do this task.\\
    \\
    \\
    \{Some Examples\}
    \\
    \\
    Your Task:\\
    QUESTION:\\
    \{Question\}\\
    \\
    RESPONSE:\\
    \{Response\}\\
    \\
    STATEMENT:\\
    \{Statement\}\\

    \bottomrule
    \end{tabularx}
    }
    \caption{Prompt for fact-level confidence estimation  \{Criterion\}, \{Question\}, \{Response\} and  \{Statement\} are placeholders.}
    \label{tab:prompt-fact-conf}
\end{table}

\subsection{Prompt for Response-Level Confidence Estimation}
The specific prompts used for response-level confidence estimation are detailed below.
\begin{table}[h!]
    \small
    {\ttfamily
    \begin{tabularx}{\linewidth}{X}
    \toprule
    Instructions:\\
    1. The following RESPONSE is the answer to the given QUESTION. \\
    2. Indicate how confident you are in the accuracy of the RESPONSE when answering the QUESTION, based on your knowledge.\\
    3. The confidence evaluation should be a value between 0 and 1 (with two decimal places retained), based on the following scoring criterion: \\
    \{Criterion\} \\
    4. Your task is to do this for the RESPONSE and QUESTION under ``Your Task''. \\
    Some examples have been provided for you to learn how to do this task.\\
    \\
    \\
    \{Some Examples\}
    \\
    \\
    Your Task:\\
    QUESTION:\\
    \{Question\}\\
    \\
    RESPONSE:\\
    \{Response\}\\

    \bottomrule
    \end{tabularx}
    }
    \caption{Prompt for response-level  confidence estimation  \{Criterion\}, \{Question\}, \{Response\} and  \{Statement\} are placeholders.}
    \label{tab:prompt-response-conf}
\end{table}

\subsection{Prompt for Factor Extraction}
The specific prompts used for factor extraction are detailed below.
\begin{table}[h!]
    \small
    {\ttfamily
    \begin{tabularx}{\linewidth}{X}
    \toprule

    Instructions:\\
    You are to read a sentence and identify the key factors within it. \\
    The task involves pinpointing the essential elements or aspects that significantly influence or characterize the situation, event, or subject described.\\
    Return the identified key factors using the format <[factor1, factor2, ...]>\\
    Some examples have been provided for you to learn how to do this task.\\
    \\
    \\
    \{Some Examples\}
    \\
    \\
    Your Task:\\
    SENTENCE:\\
    \{Sentence\}\\

    \bottomrule
    \end{tabularx}
    }
    \caption{Prompt for factor extraction  \{Sentence\} is placeholders.}
    \label{tab:prompt-response-conf}
\end{table}

\subsection{Prompt for Fact Correction}
The specific prompts used for fact correction are detailed below.
\begin{table}[h!]
    \small
    {\ttfamily
    \begin{tabularx}{\linewidth}{X}
    \toprule

    Instructions:\\
    You have been provided with a sentence and some reference knowledge. \\
    The sentence has been analyzed, and its factors have been identified. \\
    However, it is acknowledged that there may be errors or inaccuracies in the identified factors.\\
    Your task is to first review the identified factors and check for any errors or inaccuracies. \\
    If there are no errors, simply return ``NoError'' to indicate that no corrections are needed. \\
    If errors are present, proceed to make the necessary corrections.\\
    Ensure that the corrections are limited to the existing factors without adding new content.\\
    Use the format <old factor -> new factor> for each correction.\\
    \\
    \\
    \{Some Examples\}
    \\
    \\
    Your Task:\\
    SENTENCE:\\
    \{Sentence\}\\
    \\
    FACTORS:\\
    \{Factor\}\\
    \\
    REFERENCE:\\
    \{Reference\}\\

    \bottomrule
    \end{tabularx}
    }
    \caption{Prompt for fact correction \{Sentence\}, \{Factor\} and  \{Reference\} are placeholders.}
    \label{tab:prompt-response-conf}
\end{table}

\subsection{GPT-4 Judgments for Self-Correction}
For the self-correction, we utilize GPT-4 for zero-shot pair-wise evaluation.  
We use \texttt{gpt-4-0314} for all our experiments. 
The specific prompts used for GPT-4 evaluation are detailed below.

\begin{table}[h!]
    \small
    {\ttfamily
    \begin{tabularx}{\linewidth}{X}
    \toprule
    You will be provided with a QUESTION, its RESPONSE, and all facts extracted from the RESPONSE under the heading "ALL FACTS". You will also be provided with a specific fact under the heading "TARGET FACT 1", which is included in ALL FACTS. Additionally, you will be given a modified version of this target fact under the heading "TARGET FACT 2". \\
    \\
    Based on your knowledge, evaluate whether the modification of the target fact is an improvement, the same, or a regression. \\

    An improvement implies: \\
    1. More accurate information. \\
    2. Greater relevance to the question.\\
    3. Minimal overlap with other facts in ALL FACTS.\\
    \\
    A regression implies: 
    1. Introduction of erroneous or inaccurate information.\\
    2. Lower relevance to the question.\\
    3. Repetition or introduction of information that is already provided with other facts in ALL FACTS.\\
    \\
    \\
    QUESTION:\\
    \{Question\}
    \\
    \\
    RESPONSE:\\
    \{Response\}
    \\
    \\
    ALL FACTS:\\
    \{All Facts\}
    \\
    \\
    TARGET FACT 1:\\
    \{Original Fact\}
    \\
    \\
    TARGET FACT 2:\\
    \{New Fact\}
    \\
    \\
    First, provide a one-sentence comparison of the two facts and explain whether you think the modification is an improvement, the same, or a regression.
    Second, on a new line, the state only "IMPROVED", "SAME", or "REGRESSED" to indicate the effectiveness of the modification. Your response should use the following format:\\
    COMPARISON:
    <one-sentence comparison and explanation>\\
    REVISION: <"IMPROVED", "SAME", or "REGRESSED">\\
    
    \bottomrule
    \end{tabularx}
    }
    \caption{Prompt for GPT-4 evaluation for the self-correction   \{Question\}, \{Response\}, \{All Facts\},  \{All Facts\}, \{Original Facts\} and \{New Fact\} are placeholders.}
    \label{tab:prompt-gpt4}
\end{table}

\section{Related Work}
The concept of confidence calibration was first introduced to nerual networks by~\cite{guo2017calibration} to prevent logits from making incorrect classifications with high probability. This concept has since been extended to NLP models~\cite{desai2020calibration, dan2021effects, hu2023uncertainty}. 
Common methods for estimating confidence scores include logit-based methods, consistency-based methods, and verbalization-based methods.
Logit-based methods \citep{DBLP:conf/icml/GuoPSW17, cheng2023prompting,kadavath2022language} assess model confidence by examining the logits predicted by the model.
Consistency-based methods \citep{DBLP:conf/iclr/0002WSLCNCZ23, DBLP:conf/iclr/KuhnGF23} rely on the principle that language models tend to produce similar outputs consistently when they are confident.
Recently, research has indicated that verbalization-based methods~\cite{tian2023just} might offer superior confidence estimation.

\label{sec:appendix}

\end{document}